\documentclass{article}
\usepackage{spconf,amsmath,graphicx}
\usepackage{amssymb}
\usepackage{soul} 
\usepackage{amsmath} 
\usepackage{booktabs} 
\usepackage{enumitem}
\usepackage{multirow}
\usepackage{xspace} 
\usepackage{microtype} 
\usepackage{CJKutf8} 
\usepackage{subcaption} 
\usepackage{tablefootnote}
\usepackage[switch]{lineno} 
\usepackage[ruled,linesnumbered]{algorithm2e}
\SetKwRepeat{Do}{do}{while}%

\usepackage{hyperref}
\hypersetup{hidelinks, colorlinks=true, allcolors=black}

\newcommand\our{\mbox{AURL}\xspace}

\newcommand\sysdstbuffer{\mbox{$BF_{sys\_DST}$}\xspace}
\newcommand\sysdpbuffer{\mbox{$BF_{sys\_DP}$}\xspace}
\newcommand\usernlubuffer{\mbox{$BF_{user\_NLU}$}\xspace}
\newcommand\userdpbuffer{\mbox{$BF_{user\_DP}$}\xspace}
\newcommand{\tabincell}[2]{\begin{tabular}{@{}#1@{}}#2\end{tabular}} 

\title{An Asynchronous Updating Reinforcement Learning Framework for Task-oriented Dialog System}
\name{Sai Zhang, Yuwei Hu, Xiaojie Wang\sthanks{Corresponding author.} and Caixia Yuan}
\address{Beijing University of Posts and Telecommunications, Beijing, China \\ \texttt{\{zs,hyw724,xjwang,yuancx\}@bupt.edu.cn}}

\begin{document}
%
\maketitle
\begin{abstract}
    Reinforcement learning has been applied to train the dialog systems in many works. Previous approaches divide the dialog system into multiple modules including DST (dialog state tracking) and DP (dialog policy), and train these modules simultaneously. However, different modules influence each other during training. The errors from DST might misguide the dialog policy, and the system action brings extra difficulties for the DST module. To alleviate this problem, we propose \textbf{A}synchronous \textbf{U}pdating \textbf{R}einforcement \textbf{L}earning framework (\our) that updates the DST module and the DP module asynchronously under a cooperative setting. Furthermore, curriculum learning is implemented to address the problem of unbalanced data distribution during reinforcement learning sampling, and multiple user models are introduced to increase the dialog diversity. Results on the public SSD-PHONE dataset show that our method achieves a compelling result with a $31.37\%$ improvement on the dialog success rate. The code is publicly available via \href{https://github.com/shunjiu/AURL}{https://github.com/shunjiu/AURL}.
\end{abstract}
\begin{keywords}
Task-oriented dialog system, multi-agent reinforcement learning, curriculum learning, user simulator
\end{keywords}
\section{Introduction}

Task-oriented dialog systems are widely employed for customer service, e.g., automatic ticket booking. A dialog system is usually composed of four modules: natural language understanding (NLU), dialog state tracking (DST), dialog policy (DP) and natural language generation (NLG) \cite{liu-lane-2018-end}. DST, which maintains the dialog state from the beginning of the dialog to the current turn, is usually trained using supervised learning (SL). DP, which decides the system action to guide the direction of the dialog, can be trained via SL \cite{zhang2020task} with labeled data or reinforcement learning (RL) \cite{takanobu-etal-2019-guided, takanobu2020multi} with a user simulator serving as a part of interacting environment.

\begin{figure}
    \centering
    \includegraphics[width=\linewidth]{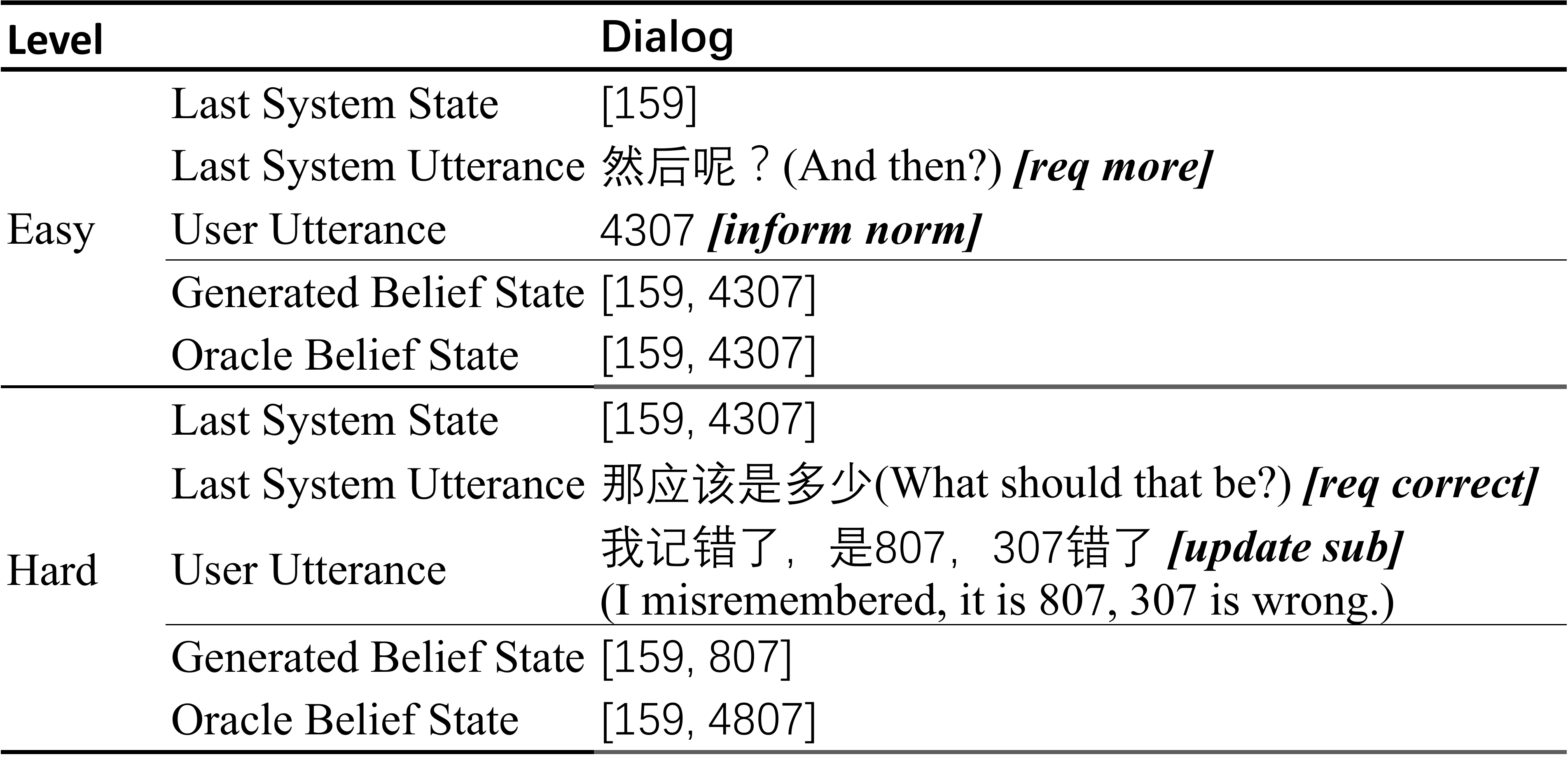}
    \caption{Example of \emph{easy} and \emph{hard} dialogs, different user actions bring different challenges for the system DST module. The user action \texttt{[inform norm]} is easy for system while \texttt{[update sub]} is hard. The errors from the DST module may misguide the decision from the system DP module. The user action is related to the system action decided by the system DP module.}
    \label{fig: dialogexample}
\end{figure}

In previous works, when training DP using RL, rule-based DST is usually applied \cite{takanobu2020multi, tang-etal-2021-high} to ignore errors from DST.  However, the DST module is unstable in real scenarios. DP should guide the dialog successfully under the influence of DST errors. \cite{lei2020cooperative, liu2021converse} train the modules with SL along with RL loss from DP. But the modules will influence each other when training simultaneously.
If DST tracks wrong slot values, the system may collect a wrong slot under the right policy, leading to the policy incorrectly learning. Meanwhile, suppose the dialog policy module often chooses actions that user can respond easily, the DST module can not be trained sufficiently and may lose the ability to understand the hard user actions. As an example, Figure~\ref{fig: dialogexample}, shows that different user actions pose different challenges for the DST module, which brings bias when training. Furthermore, the DP is usually trained by interacting with a predefined user simulator~\cite{tang-etal-2021-high}, though stochastic yet monotonous. As observed in \cite{lei2020cooperative}, different users bring diverse dialogue states, thus the dialog policy can be trained sufficiently.   

To overcome these problems, we first propose a new updating reinforcement learning framework for dialog systems, where DST and DP modules are both trainable cooperatively, but each is updated asynchronously with different updating frequency. We then construct a \textbf{M}ulti-\textbf{U}ser \textbf{R}einforcement \textbf{L}earning (MURL) with the \our framework, where multiple user models are used to interact with one dialog system to get more diverse dialog strategies. In addition, we use curriculum learning \cite{bengio2009curriculum} to boost the DST learning by gradually increasing the complexity of the data samples used during the training process.

To better verify our approach, we conduct experiments on the SSD-PHONE dataset \cite{zhang-etal-2022-slot}, a large action space dialog dataset with diversity phenomena. A hierarchical neural network dialog system and a user model are built. The results demonstrate that the system trained using the proposed framework achieves a new start-of-the-art in online test. In summary, the contributions of this paper are as follows.

\begin{itemize}
    \item We propose a novel asynchronous updating reinforcement learning framework for dialog systems which trains DP and DST modules asynchronously in a cooperative setting, and applies curriculum learning to solve the bias during training.
    
    \item We propose a novel reinforcement learning framework for training one dialog agent with multiple user models, which increases the dialog diversity and sufficiently promotes dialog policy learning. 

    \item We conduct experiments on the real-world SSD-PHONE dataset. Results show the superiority of our approach to several strong baselines. Significantly, it increases by $31.37\%$ on dialog success rate than the current SOTA.

\end{itemize}

\section{Method}

\subsection{Asynchronous Updating Reinforcement Learning Framework}

\subsubsection{Framework}

\begin{figure}
    \centering
    \includegraphics[width=\linewidth]{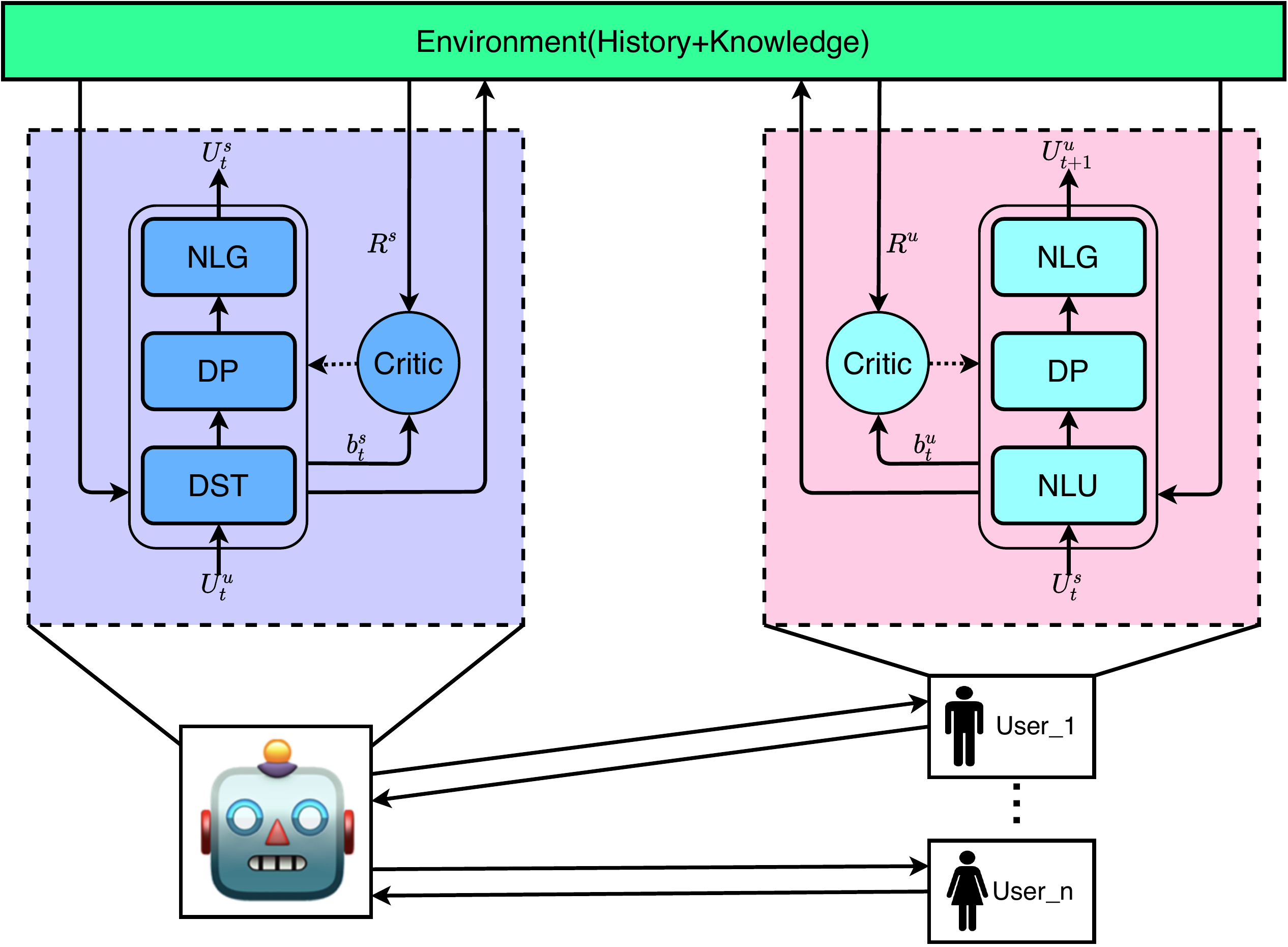}
    \caption{\our framework on multi-agent learning. System and users interact with each other via written language. The system DST module and DP module are updated asynchronously. 
    }
    \label{fig: framework}
\end{figure}

Figure~\ref{fig: framework} shows the general architecture and information flow of our framework, composed of one system agent and $N$ user agents. System and Users communicate with each other via script language. After completing one dialog, the inputs, outputs, labels or rewards for each module are kept in replay buffer \userdpbuffer, \usernlubuffer, \sysdpbuffer and \sysdstbuffer respectively. We update each module asynchronously using experience replay\cite{schaul2015prioritized}.

\noindent \textbf{System DST} produces $\phi$ that updates the dialog belief state of the current turn.  Inspired by TRADE~\cite{wu-etal-2019-transferable}, we use two encoders to encode the dialog history $[U^s_{t-1};U^u_t]$ and last system belief state $BS^s_{t-1}$ respectively, where $U^s_{t-1}$ is the system response at $t-1$ turn, $U^u_t$ is the user utterance at current turn $t$. A state generator~\cite{heck-EtAl:2020:sigdial} is applied to generate the current turn's belief state $BS^s_t$. User action $\widehat{a^u_{t}}$ is obtained from a multi-layer perceptron with input the hidden states of two encoders. System DST is formulated as:
\begin{equation}
    (BS^s_t,\widehat{a^u_t})=\phi(BS^s_{t-1},[U^s_{t-1};U^u_t]).
\end{equation}

\noindent \textbf{System DP} produces $\pi$ that decides the current turn's system action $a^s_t$ and system slot $s^s_t$ according to the dialog state. The dialog state at dialog turn $t$ is the concatenation of (1) the system action at last turn $a^s_{t-1}$, (2) the belief state at current turn $BS^s_t$, (3) the hidden state from dialog history encoder $H^{ctx}_t$, (4) the query results $q_t$ from \texttt{DB}. System DP is formulated as:
\begin{equation}
    (a^s_t,s^s_t)=\pi(a^s_{t-1};BS^s_t;H^{ctx}_t;q_t).
\end{equation}

\noindent \textbf{User NLU} yields $\eta$ that understands the system action $\widehat{a^s_t}$ and system slot $\widehat{s^s_t}$ according to system utterance $U^s_t$ and user's goal value $G$. User NLU is formulated as:
\begin{equation}
    (\widehat{a^s_t},\widehat{s^s_t})=\eta(G;U^s_t).
\end{equation}

\noindent \textbf{User DP} yields $\mu$ that decides user action $a^u_{t+1}$ and user slot $s^u_{t+1}$ to interact with system agent. And then, a state vector $BS^u_{t+1}$ records each slot or each part of slot if provided or need to be updated. Each slot uses a Likert scale of 0-2, which respectively represent \emph{not provided}, \emph{provided} and \emph{need updated}. To model errors brought by Automatic Speech Recognition(ASR), we randomly replace the slot with a similar one. The input to the user policy module is the concatenation of (1) the user action at last turn $a^u_t$, (2) the system action $\widehat{a^s_t}$, (3) the system slot $\widehat{s^s_t}$ and (4) the user state vector at last turn $BS^u_{t}$. User policy is formulated as:
\begin{equation}
    (a^u_{t+1},s^u_{t+1})=\mu(a^u_t;\widehat{a^s_t};\widehat{s^s_t};BS^u_t).
\end{equation}

To mitigate the errors entangling of DST and DP, we train them asynchronously with different updating frequency. We train both modules by experience replay which samples training examples when the experience pools are full. The DP is updated more frequently by examples from a smaller experience pools, while the DST is updated slowly by examples from a bigger one. In so doing, DP can optimize its policy quickly in a relative stable environments, while DST can learn with the latest DP module. 

\subsubsection{Curriculum Learning}

During reinforcement learning, training examples are randomly sampled from experience pools.  To imitate how humans learn, we introduce curriculum learning to schedule the training process. After pretrained on the offline dialog dataset, the DST model is evaluated on test dataset and obtains the joint accuracy of user action understanding. The joint accuracy is then used as difficulty measurer~\cite{XinWang2020ASO} to split the data in \sysdstbuffer into \emph{easy}, \emph{middle} and \emph{hard} levels.

Firstly, we train the DST module using \emph{easy}, \emph{middle} and \emph{hard} levels of data in order. Secondly, using \emph{middle} and \emph{hard} levels since the \emph{easy} data takes account of $75\%$ of the whole data and the model has learned well on them. Thirdly, just using \emph{hard} data. At last, all levels of data are exploited in order to review. 

\subsubsection{MURL}

Multiple user models, which are pretrained on the dialog corpus, are used to train one dialog system. The system interacts with each user in order after completing one dialog session. The users, trained independently of each other, will have different personalities during the RL training due to the random number. For the same system utterance, different users may use different actions to respond, which brings more dialog diversities to promote the system dialog policy learning. 

\subsection{Reward}

Reward is essential for reinforcement learning to guide policy learning. The roles of the system and the user are different. System should complete the dialog successfully in shorter turns for task-oriented dialog task. c System and Users communicate cooperatively in our setting to accomplish the dialog. The reward settings for each role are shown below.

For system reward $R^s$, it consists of (1) dialog success reward and dialog failed penalty at the end of the dialog; (2) a minor dialog length penalty at each turn; (3) action and slot mismatch penalty in order to avoid the system confirming empty slot value, and so on; (4) few inappropriate system actions penalty based on user action.

Different from the system, user policy should be diverse and suitable to prompt the system policy learning. User reward $R^u$ is similar to system reward. We remove the length of dialog turns penalty for user because we cannot restrict users from ending the dialog quickly.

\begin{algorithm}[t]
	\caption{\our Framework with one user.}
	\label{alg:algorithm1}
	\KwIn{Dialog corpus $D$; system model and user model; system DST buffer size $n$}
	\KwOut{Trained system model.}
	\BlankLine
	Initialize weights $\phi$, $\pi$, $\eta$, $\mu$, $V^s$, $V^u$ randomly;
	
	Pretrain $\phi$, $\pi$, $\eta$, $\mu$ on dialog corpus $D$ using SL.
	
	\ForEach{\emph{train epoch}}{
	    \ForEach{step}{
	    Initialize user goal value $G$, user state and system state.
	    
	    System gives the utterance $U^s_0$ at the first turn.
	    
	    \Repeat{the dialog ends according to T}{
	    
	    User understands system utterance $U^s$, samples action and slot using $\eta$, $\mu$, gives response $U^u$.
	    
	    System updates its dialog state $BS^s$ according to the user response using $\phi$, and then samples action and slot using $\pi$, gives response $U^s$.
	    
	    Get terminal signal T according to $BS^s$, $G$ and the dialog length.
	    
	    Observe rewards $R^s$ and $R^u$.
	    
	    Four replay buffers record inputs, outputs, labels or rewards for each module.
	    }
	    }
		Update two critic networks, two dialog policy modules and user NLU module. Clear three buffers.
		
		\If{$|\sysdstbuffer|$ equals $n$}{
		
		Update system  DST module using curriculum learning. Clear $buffer_{system\_DST}$
		}
	}
\end{algorithm}

\subsection{Optimization}

Algorithm~\ref{alg:algorithm1} shows the entire \our algorithm under one user setting. For the system DST module and user NLU module, we can get the label outputs when interacting with each other. So we use cross-entropy loss to update both modules.

Advantage actor-critic (A2C) algorithm is used to optimize both policy modules. For each role, a critic network $V$ is applied to evaluate the state value. The critic networks aim to minimize the following loss functions:
\begin{align}
    L^s_V=(R^s+\gamma V^s(b^s_{t+1})-V^s(b^s_t))^2 , \\
    L^u_V=(R^u+\gamma V^u(b^u_{t+1})-V^u(b^u_t))^2 ,
\end{align}
where  $b^s_t = [a^s_{t-1};BS^s_t;H^{history}_t;q_t] $, $b^u_t = [a^u_t;a^s_t;s^s_t;BS^u_t]$.

The actor network (policy) aims to maximize the returns. Advantage function $A(a,s;b_t)=R+\gamma V(b_{t+1})-V(b_t)$ evaluates if the chosen action and slot are better. The loss functions for policy modules are below:
\begin{align}
    L^s_{\pi} = A(a^s;s^s,b^s)(\log_{\pi}(a^s|b^s)+\log_{\pi}(s^s|b^s)),\\
    L^u_{\mu} = A(a^u;s^u,b^u)(\log_{\mu}(a^u|b^u)+\log_{\mu}(s^u|b^u)).
\end{align}

Besides, different sizes of replay buffers are applied to update the system DST and other modules asynchronously.

\section{Experiments}

\subsection{Experimental Setup}

\noindent \textbf{Dataset.} 
SSD-PHONE \cite{zhang-etal-2022-slot} is a real-world task-oriented dialog corpus that contains $30$ actions, $11,000$ dialog sessions, $3,135$ different dialog paths and plenty of diversity phenomena. The corpus also provides a wealth of annotation information. To verify the ability of reinforcement learning in more challenging scenario, we further expand the number of dialog actions to $41$ according to the diversity phenomena, including $16$ system actions and $25$ user actions.

\noindent \textbf{Evaluation Metrics.} 
We evaluate model performances by online interacting with the FSA-based user simulator provided by \cite{zhang-etal-2022-slot}. After interacting three times with $1,000$ dialogs each, we calculate the following metrics. \textbf{Dialog succ} is the main metric. A dialog is successful if and only if the slot values collected by the system is equal to the user goal value within limited turns. \textbf{Avg turn} shows the average turn number of successful dialogs. \textbf{Avg reward} is the average of the system reward for each dialog. \textbf{DST acc} means whether the slot values are correctly collected at each turn. \textbf{Avg time} measures the average response time interacting with users.



\noindent \textbf{Implementation Details.}
The size of reply buffer is $6$, $6$, $6$, $3k$ for \userdpbuffer, \usernlubuffer, \sysdpbuffer and \sysdstbuffer respectively. During RL training, $100k$ epochs and $6$ dialogs in each epoch are trained. In terms of reward design, the rewards for system and user are both set to $2.0$ if dialog succeeds, otherwise, both set to $-1.0$. Penalties of other types for system are set to $-0.05$ and for user are all set to $-0.02$. $\gamma$ is set to $0.99$.

\subsection{Baselines}
In addition to the baselines in \cite{zhang-etal-2022-slot}, we compare \our with different methods. \textbf{DAMD}\cite{zhang2020task} trains dialog policy using SL, considering that a dialog state may correspond to many system actions. 
\textbf{SL} is the system model pretrained on the SSD-PHONE dataset first. 
The following baselines all use the pretrained system model and user model. We conduct experiments using REINFORCE \cite{williams1992simple}, A2C, asynchronous advantage actor-critic \cite{10.5555/3045390.3045594} and proximal policy optimization \cite{schulman2017proximal} algorithm, finally adopt A2C as our learning algorithm since A2C has a more stable learning curve.
\textbf{RL-fixed\_DST} is the traditional RL setting, in which just the system policy module and the user policy module are updated.
\textbf{RL-train\_DST} denotes that all modules are trained simultaneously.

\begin{table}[t]
\small
\begin{tabular}{c|cccccc} \hline
Model & \tabincell{c}{Dialog \\ succ} & \tabincell{c}{Avg \\ turn} & \tabincell{c}{Avg \\ reward} & \tabincell{c}{DST \\ acc} & \tabincell{c}{AVG \\ time} \\ \hline
TRADE*  & $30.45$   &  $9.77$      &  -   & - & $111$ \\
DAMD & $46.40$ & $6.52$ & - & - & $489$ \\
UBAR & $57.70$ & $11.39$ & - & - & $376$ \\
SimpleTOD  & $63.20$  & $8.18$ & - & - & - \\ 
SL & $\textbf{75.73}$ &  $7.35$ & $1.19$ & $71.12$ & $\textbf{28}$ \\ \hline
RL-fixed\_DST & $77.23$  & $6.93$ & $1.28$ & $66.78$ & -\\
RL-train\_DST$^{\star}$ & $71.00$  & $\textbf{5.94}$ & $1.11$ & $75.58$ & - \\
\our & $ 84.13 $ & $7.36$ & $ 1.47 $ & $ 70.92$ & - \\
\our -1v2 & $\textbf{94.57}$ & $8.32$ & $\textbf{1.82}$ & $\textbf{81.11}$ & - \\

\hline

\end{tabular}
\caption{Results of different models on interaction with the FSA-based user simulator. \our -1v2 denotes  using two user models to train the system model. Only one user model is applied under the other RL settings. $\star$ is the ablation study.}
\label{tbl:onlinetest}
\end{table}

\subsection{Results and Analysis}

The online evaluation results of each model are summarised in Table~\ref{tbl:onlinetest}. Among the supervised learning settings, the proposed model performs the best, with an $11.40\%$ improvement over GPT2 \cite{radford2019language} based models (SimpleTOD \cite{2020-simpleTOD}). Furthermore, our model is lighter than other baselines, and with a shorter response time per turn, which gives responses more quickly when interacting with users when deployed on a real dialing platform. 

Under the RL setting, the comparison between RL-fix\_DST and RL-train\_DST indicates the bias in the simultaneous training method. When using one user model to train the dialog system, \our reaches a higher dialog success than SL with an $8.40\%$ improvement. With the aid of the asynchronous updating and curriculum learning, the DP module even makes right decision with the DST errors.

When using two user models to train one dialog system, the dialog success rate is improved from $84.13\%$ to $94.57\%$. Besides, the accuracy of the DST module reaches $81.11\%$. More user models bring more different dialog paths like various humans in reality, even though they are initialized with the same pretrained parameters. The dialog system is trained more adequately than using only one user model. 

\section{Conclusions}

In this paper, we proposed an asynchronous updating reinforcement learning framework for DST and DP modules of task-oriented dialog system. We conducted multi-agent reinforcement learning in asynchronous updating framework to train both models. With the benefit of curriculum learning and multiple user models training, our approach achieves a new SOTA, with a $31.37\%$ improvement over original GPT2-based models on the online test. In the future, more work will be done under the MURL setting, especially for introducing more user models to train the system agent.

\section{Acknowledgments}

We thank the anonymous reviewers for their insightful comments. The research is supported by the Major Research Plan of National Natural Science Foundation of China (Grant No.92067202).
\clearpage
\bibliographystyle{unsrt}
\bibliography{ref}
\end{document}